\documentclass[runningheads]{llncs}
\usepackage{graphicx}
\usepackage{mathtools}
\usepackage[utf8]{inputenc}
\usepackage[vietnamese,english]{babel}
\usepackage{pgfplots}
\usepackage{tikz}
\usepackage{amsmath,amssymb}
\usetikzlibrary{patterns}
\pgfplotsset{compat=1.5}
\usepackage{subcaption}
\usepackage[ruled,lined,boxed,commentsnumbered]{algorithm2e}

\begin{document}
\title{ImageNet Challenging Classification with the Raspberry Pi: An Incremental Local Stochastic Gradient Descent Algorithm}
\titlerunning{ImageNet Classification with the RPi: An Incremental Local SGD}
\author{Thanh-Nghi Do\inst{1,2}}
\authorrunning{T-N. Do}
\institute{College of Information Technology \\
			   Can Tho University, 92000-Cantho, Vietnam \\
			   \and
			   UMI UMMISCO 209 (IRD/UPMC) \\
			   Sorbonne University, Pierre and Marie Curie University - Paris 6, France \\
               \email{dtnghi@ctu.edu.vn}}
\maketitle              % typeset the header of the contribution

\begin{abstract}
With rising powerful, low-cost embedded devices, the edge computing has become an increasingly popular choice. In this paper, we propose a new incremental local stochastic gradient descent (SGD) tailored on the Raspberry Pi to deal with large ImageNet ILSVRC 2010 dataset having 1,261,405 images with 1,000 classes. The local SGD splits the data block into $k$ partitions using $k$means algorithm and then it learns in the parallel way SGD models in each data partition to classify the data locally. The incremental local SGD sequentially loads small data blocks of the training dataset to learn local SGD models. The numerical test results on Imagenet dataset show that our incremental local SGD algorithm with the Raspberry Pi 4 is faster and more accurate than the state-of-the-art linear SVM run on a PC Intel(R) Core i7-4790 CPU, 3.6 GHz, 4 cores. 

\end{abstract}
\setcounter{footnote}{0}

\section{Introduction}
\label{sec:introduction}
The efficient image classification algorithms allow to find what we are looking for in very large amount of images produced by internet users. The image classification task automatically categorizes the image into one of predefined classes. It consists of two key stages: the feature extraction and the machine learning scheme. The classical approaches \cite{bosch_2006,do_vjcs_sgd14,do_iccsama_mclr15,do_fdse_16,do_mta15,LiPT_05,SivicZ_03,PerronninSL10,Wu_CVPR_2012}  proposed to use popular handcrafted features such as the scale-invariant feature transform (SIFT \cite{lowe_1999,lowe_2004}), the bag-of-words model (BoW) and then to train Support Vector Machines (SVM \cite{vapnik_SVM_1995}) to classify images. Recent convolutional neural networks (CNN \cite{lecun_lenet_1998}), deep neural networks including VGG19 \cite{Simonyan_VGG_14}, ResNet50 \cite{HeZRS_ResNet_15}, Inception v3 \cite{SzegedyVISW_inception_15}, Xception \cite{Chollet_xception_16} aim to learn visual features from images and the classifier in an unified algorithm to efficiently classify images. These deep networks achieve the prediction correctness more over 70\% for ImageNet challenging dataset \cite{DengBLF10,Deng_CVPR_2009}.

In this paper, we use the pre-trained deep learning network Inception v3 \cite{SzegedyVISW_inception_15} to extract invariant features from images. Followed which, we develop the new incremental local SGD algorithm tailored on the Raspberry Pi for classifying very large ImageNet ILSVRC 2010 dataset. To overcome the main memory limit, the incremental local SGD sequentially loads small data blocks of large trainset to learn local SGD models. The local SGD uses $k$means algorithm \cite{macqueen_1967} to split the training data block into $k$ data partitions and then it learns in the parallel way SGD models in each data partition to classify the data locally. The numerical test results on ImageNet dataset show that our incremental local SGD algorithm on the Raspberry Pi 4 (Broadcom BCM2711, Quad core Cortex-A72 (ARM v8) 64-bit SoC @ 1.5GHz, 4GB RAM) is faster and more accurate than the state-of-the-art linear SVM such as LIBLINEAR \cite{fan_liblinear_2008} run on a PC (Intel(R) Core i7-4790 CPU, 3.6 GHz, 4 cores and 32 GB RAM). The incremental local SGD classifies ImageNet dataset having 1,261,405 images in 2048 deep features into 1,000 classes with an accuracy of 75.61\% in 2 hours and 9.48 minutes using the Raspberry Pi 4.

The remainder of this paper is organized as follows. Section \ref{sec:inc-lsgd} briefly presents the incremental local SGD algorithm. Section~\ref{sec:evaluation} shows the experimental results before conclusions and future works presented in section~\ref{sec:conclusion}.
\section{Incremental local stochastic gradient descent}
\label{sec:inc-lsgd}
Let us consider a classification task with the dataset $D=[X, Y]$ consisting of $m$ datapoints $X = \lbrace x_1, x_2, \ldots, x_m \rbrace$ in the $n$-dimensional input space $R^n$, having corresponding labels $Y = \lbrace y_1, y_2, \ldots, y_m \rbrace$ being $\lbrace c_1, c_2, \ldots, c_p \rbrace$. 

\subsection{Stochastic gradient descent}

The stochastic gradient descent (SGD) algorithm tries to find $p$ separating planes for $p$ classes (denoted by normal vectors $w_1, w_2, \dots, w_p \in R^n$) in which the plane $w_i$ separates the class $c_i$ from the rest. This is accomplished through the unconstrained problem (\ref{uqp1}).

\begin{equation}
\label{uqp1}
\min\ \Psi(w_p, [X, Y]) = \frac{\lambda}{2}\|w_p\|^{2} + \frac{1}{m} \sum_{i=1}^m{L(w_p, [x_i, y_i])}
\end{equation}

\noindent{where the errors are measured by $L(w_p, [x_i, y_i]) = max\{0, 1 - y_i(w_p.x_i)\}$ and a positive constant $\lambda$ is to control the regularization strength ($\|w_p\|^{2}$).}

Studies in \cite{Bottou_SGD08,Shwartz_PEGASOS07} illustrate that the (SGD) algorithm solves the unconstrained problem (\ref{uqp1}) by updating $w$ on $T$ epochs with a learning rate $\eta$. For each epoch $t$, the SGD uses a single datapoint ($x_i, y_i$) randomly in the mini batch $B_i$ to compute the sub-gradient $\nabla_t  \Psi(w_p, [x_i, y_i])$ and update $w_{p}$ as follows:

\begin{eqnarray}
\label{update1}
w_{p}  = w_{p} - \eta \nabla_t  \Psi(w_{p}, [x_i, y_i]) 
\end{eqnarray}

The SGD is a simple yet efficient algorithm for large-scale learning due to the computational complexity corresponding to $O(mnp)$ (linear in the number of training datapoints $m$). 

In recent last years, it rises powerful, low-cost embedded devices. For example, the Raspberry Pi 4 (Broadcom BCM2711, Quad core Cortex-A72 (ARM v8) 64-bit SoC @ 1.5GHz, 4GB RAM) is only 55 USD. This leads an increasingly popular choice for machine learning and IoT projects, as illustration in \cite{Kouli_19,Kulkarni_20,Kurniawan_21,Norris_20}, image classification on IoT edge devices \cite{MagidPD20}, MobileNet family tailored for Raspberry Pi \cite{GlegolaKP21}, running AlexNet on Raspberry Pi \cite{Iodice_2018}. 

Nevertheless, it is intractable to train the SGD model with the Raspberry Pi for ImageNet challenging ILSVRC 2010 problem having 1,261,405 images with 1,000 classes since it requires at least 16 GB RAM for loading the training dataset and the high computational cost. Our investigation aims to reduce the training time and the required main memory of the SGD algorithm, being tailored on the Raspberry Pi.

\subsection{Incremental local stochastic gradient descent}

As illustration in Fig. \ref{fig:ksgd}, our proposed local SGD (denoted by $k$SGD) uses $k$means  \cite{macqueen_1967} to divide the training data block into $k$ partitions  and then it learns local SGD models in data partitions in parallel way on multi-core computers. 

\begin{figure}
\resizebox{1.\textwidth}{!}{

\begin{tabular}{c}
\begin{subfigure}{1.0\textwidth}
\centering
\begin{tikzpicture}

\node[draw] (ts) at (-2.2,5.5) {Training dataset $D$};

\node[draw] (km) at (-2.2,3.5) {Partition $D$ into $3$ clusters with $k$-means};

\draw [->] (ts) -- (km);

\node[circle,draw] (c1) at (-6.7,1) {$D_1$};
\node[circle,draw] (c2) at (-2.2,1) {$D_2$};
\node[circle,draw] (c3) at (2.2,1) {$D_3$};
\draw [->] (km) -- (c1);
\draw [->] (km) -- (c2);
\draw [->] (km) -- (c3);

\node[draw] (sgd1) at (-6.7,-0.8) {$c_1, lSGD_1 = SGD(D_1, \theta)$};
\node[draw] (sgd2) at (-2.2,-0.8) {$c_2, lSGD_2 = SGD(D_2, \theta)$};
\node[draw] (sgd3) at (2.2,-0.8) {$c_3, lSGD_3 = SGD(D_3, \theta)$};
\draw [->] (c1) -- (sgd1);
\draw [->] (c2) -- (sgd2);
\draw [->] (c3) -- (sgd3);

\end{tikzpicture}
\end{subfigure} \\
\\
\\
\begin{subfigure}{1.0\textwidth}
\centering
\begin{tikzpicture}
\begin{axis}[
scatter/classes={
a={mark=square,black},%
b={mark=x,black},%
c={mark=o,draw=black},
d={mark=triangle,draw=black}}]
\addplot[scatter,only marks,
scatter src=explicit symbolic]
		coordinates {
(0.686000,0.110000) [a]
(0.638000,0.134000) [a]
(0.560000,0.164000) [a]
(0.608000,0.204000) [a]
(0.670000,0.246000) [a]
(0.720000,0.164000) [a]
(0.680000,0.196000) [a]
(0.582000,0.270000) [b]
(0.596000,0.328000) [b]
(0.668000,0.298000) [b]
(0.750000,0.274000) [b]
(0.726000,0.346000) [b]
(0.760000,0.226000) [b]
(0.524000,0.412000) [d]
(0.502000,0.454000) [d]
(0.432000,0.484000) [d]
(0.426000,0.540000) [d]
(0.540000,0.528000) [d]
(0.574000,0.492000) [d]
(0.482000,0.340000) [b]
(0.478000,0.418000) [b]
(0.416000,0.448000) [b]
(0.340000,0.442000) [b]
(0.394000,0.388000) [b]
(0.312000,0.148000) [c]
(0.270000,0.116000) [c]
(0.224000,0.084000) [c]
(0.330000,0.062000) [c]
(0.258000,0.178000) [c]
(0.382000,0.110000) [b]
(0.396000,0.166000) [b]
(0.330000,0.236000) [b]
(0.254000,0.244000) [b]
(0.324000,0.194000) [b]
(0.418000,0.244000) [b]

		};

\draw[color=green,dashed] (axis cs:0.32,0.17) circle (130);
\draw[color=green,dashed] (axis cs:0.46,0.455) circle (130);
\draw[color=green,dashed] (axis cs:0.665,0.233) circle (130);

%lsvm1	
\addplot[smooth,color=red,dashed] 
		coordinates {
(0.228000,0.261000) [1]
(0.405000,0.069000) [1]
		}; 	

%lsvm2					
\addplot[smooth,color=red,dashed] 
		coordinates {
(0.350000,0.512000) [1]
(0.563000,0.372000) [1]
		}; 	

%lsvm3		
\addplot[smooth,color=red,dashed] 
		coordinates {
(0.540000,0.250000) [1]
(0.788000,0.190000) [1]
		}; 	

\node at (axis cs:0.32,0.17) {$c_1$};
\node at (axis cs:0.46,0.455) {$c_2$};
\node at (axis cs:0.665,0.233) {$c_3$};

\node at (axis cs:0.31,0.27) {$D_1$};
\node at (axis cs:0.49,0.55) {$D_2$};
\node at (axis cs:0.65,0.33) {$D_3$};

\node at (axis cs:0.46,0.06) {$lSGD_1$};
\node at (axis cs:0.28,0.51) {$lSGD_2$};
\node at (axis cs:0.49,0.27) {$lSGD_3$};

\end{axis}

\end{tikzpicture}
\end{subfigure}
\end{tabular}
} % resize
\caption{Three local SGD model ($k$SGD)}
\label{fig:ksgd}
\end{figure}

%Our proposed local SGD (denoted by $k$SGD) splits the training data block into $k$ partitions using $k$means algorithm \cite{macqueen_1967} and then it learns local SGD models in data partitions in parallel way on multi-core computers.

\IncMargin{1em}
\begin{algorithm}
\LinesNumbered
\SetKwInOut{Input}{input}
\SetKwInOut{Output}{output}
\Input{
\\
training dataset $D$ \\
number of local models $k$ \\
parameters of SGD $\theta$ \\
}
\Output{
\\
$k$SGD-model ($k$ local SGD models)
}
\BlankLine
\Begin{
/*$k$-means performs the data clustering on $D$;*/ \\
creating $k$ clusters denoted by $D_1, D_2, \dots, D_k$ and \\
their corresponding centers $c_1, c_2, \dots, c_k$ \\
\#pragma omp parallel for \\
\For{$i\leftarrow 1$ \KwTo $k$}{
/*learning a local SGD model from $D_i$;*/ \\
$SGD_i = SGD(D_i, \theta)$
}

return $k$SGD-model = \{$(c_1, SGD_1), (c_2, SGD_2), \dots, (c_k, SGD_k)$\} 
}
\caption{$k$ local SGD algorithm ($k$SGD)}
\label{alg:localsgd}
\end{algorithm}\DecMargin{1em}

Algorithms \ref{alg:localsgd} and \ref{alg:pred-localsgd} describe the $k$ local SGD learning stage and prediction, respectively. 

The $k$ local SGD algorithm not only reduces the training complexity of the full SGD and but also allows to parallelize the training task of $k$ local SGD models on multi-core computers.

Let us to illustrate the complexity of the $k$ local SGD algorithm. Splitting the full training dataset with $m$ datapoints in $n$ dimensions and $p$ classes into $k$ balanced clusters leads the cluster size being about $\frac{m}{k}$ and the number of classes in a cluster scaling $\frac{p\omega}{k} < p$. The training complexity of a local SGD \footnote{It must be noted that the complexity does not include the minibatch $k$-means \cite{minibatchkmeans_10} used to partition the full dataset.} is $O(\frac{m}{k}n\frac{p\omega}{k})$. Therefore, the complexity of parallel training $k$ local SGD models on a $P$-core processor is $O(\frac{m}{Pk}\omega np)$. This illustrates that parallel learning $k$ local SGD models is $\frac{Pk}{\omega}$ times faster than the global SGD training ($O(mnp)$).  

Studies in \cite{bottou_local_1992,do_16,vapnik_local_1993} point out the trade-off between the capacity of the local learning algorithm and the complexity. The large value of $k$ reduces significant training time of $k$SGD and making a very low generalization capacity. The small value of $k$ improves the generalization capacity but also increasing the training time.

\IncMargin{1em}
\begin{algorithm}
\LinesNumbered
\SetKwInOut{Input}{input}
\SetKwInOut{Output}{output}
\Input{
\\
a new datapoint $x$ \\
$k$SGD-model = \{$(c_1, SGD_1), (c_2, SGD_2), \dots, (c_k, SGD_k)$\} \\
}
\Output{
\\
predicted class $\hat{y}$
}
\BlankLine
\Begin{
/* find the closest cluster based on the distance between $x$ and cluster centers $c_1, c_2, \dots, c_k$ */ \\

$c_{NN} = \arg\min_c\ distance(x, c)$ \\

/* the class of $x$ is predicted by the local SGD model $SGD_{NN}$ corresponding to $c_{NN}$ */ \\
$\hat{y} = predict(x, SGD_{NN})$

return predicted class $\hat{y}$
}
\caption{Prediction of a new individual $x$ with $k$SGD model}
\label{alg:pred-localsgd}
\end{algorithm}\DecMargin{1em} 

To overcome the main memory limit of the Raspberry Pi, we propose to train local SGD models in the incremental fashion. The full training set $D$ is split into $T$ small blocks $\{D_1, D_2, \dots, D_T\}$. The incremental local SGD (denoted by Inc-$k$SGD in Algorithm \ref{alg:inc-localsgd}) sequentially loads data block $D_t$ to learn local SGD models. The prediction of a new datapoint $x$ (in algorithm \ref{alg:pred-inc-localsgd}) is the majority vote among classification results {$\hat{y}_1, \hat{y}_2, \dots, \hat{y}_T$\} obtained by $T$ $k$SGD models.

The incremental local SGD can be explained by training an ensemble of local SGD models.  

\IncMargin{1em}
\begin{algorithm}
\LinesNumbered
\SetKwInOut{Input}{input}
\SetKwInOut{Output}{output}
\Input{
\\
training dataset $D = \{D_1, D_2, \dots, D_T\}$\\
number of local models $k$ \\
parameters of SGD $\theta$ \\
}
\Output{
\\
inc-$k$SGD-model (ensemble of $k$SGD models)
}
\BlankLine
\Begin{
\For{$t\leftarrow 1$ \KwTo $T$}{
Loading data block $D_t$ \\
$kSGD_t = kSGD(D_t, k, \theta)$
}

return inc-$k$SGD-model = \{$kSGD_1, kSGD_2, \dots, kSGD_T$\} 
}
\caption{Incremental $k$ local SGD (Inc-$k$SGD)}
\label{alg:inc-localsgd}
\end{algorithm}\DecMargin{1em}

\IncMargin{1em}
\begin{algorithm}
\LinesNumbered
\SetKwInOut{Input}{input}
\SetKwInOut{Output}{output}
\Input{
\\
a new datapoint $x$ \\
inc-$k$SGD-model = \{$kSGD_1, kSGD_2, \dots, kSGD_T$\} \\
}
\Output{
\\
predicted class $\hat{y}$
}
\BlankLine
\Begin{

\For{$t\leftarrow 1$ \KwTo $T$}{
$\hat{y}_t = predict(x, kSGD_t)$
}

$\hat{y}$ = majority-vote\{$\hat{y}_1, \hat{y}_2, \dots, \hat{y}_T$\}

return predicted class $\hat{y}$
}
\caption{Prediction of a new individual $x$ with inc-$k$SGD model}
\label{alg:pred-inc-localsgd}
\end{algorithm}\DecMargin{1em}

\section{Experimental results}
\label{sec:evaluation}

We are interested in the assessment of the incremental local SGD (Inc-$k$SGD) algorithm with the Raspberry Pi for handling ImageNet challenging dataset. Therefore, it needs to evaluate the performance in terms of training time and classification correctness.

\subsection{Software programs}

We implemented the Inc-$k$SGD in Python using library Scikit-learn \cite{scikit-learn_2011}. The full SGD algorithm is already implemented in Scikit-learn. We would like to compare with the best state-of-the-art linear SVM algorithm, LIBLINEAR \cite{fan_liblinear_2008} implemented in C/C++ (the parallel version on multi-core computers with OpenMP \cite{OpenMP_2008}) and the full SGD algorithm.

Our Inc-$k$SGD trains the ensemble of local SGD models with the Raspberry Pi 4 (RPi4) Raspbian Bulleyes, Broadcom BCM2711, Quad core Cortex-A72 (ARM v8) 64-bit SoC @ 1.5GHz, 4GB RAM. LIBLINEAR, full SGD learn classification models on a machine (PC) Linux Fedora 32, Intel(R) Core i7-4790 CPU, 3.6 GHz, 4 cores and 32 GB main memory.

\subsection{ImageNet challenging dataset}

Experimental results are evaluated on ImageNet challenging ILSVRC2010 dataset \cite{DengBLF10,Deng_CVPR_2009} with 1,261,405 images and 1,000 classes which is the most popular visual classification benchmark \cite{Chollet_xception_16,DengBLF10,Deng_CVPR_2009,do_vjcs_sgd14,dopsvm_21,do_iccsama_mclr15,dosgd_21,do_fdse_16,do_mta15,HeZRS_ResNet_15,Simonyan_VGG_14,SzegedyVISW_inception_15,Wu_CVPR_2012}.
%We uses pre-trained Inception v3 \cite{SzegedyVISW_inception_15} to extract 2,048 invariant features from images. 

We propose to use pre-trained Inception v3 \cite{SzegedyVISW_inception_15} described in Fig. \ref{fig:inc} to extract 2,048 invariant features from images (getting the last \textbf{AvgPool} layer). 

\begin{figure}
\center
  \includegraphics[height=3.2cm]{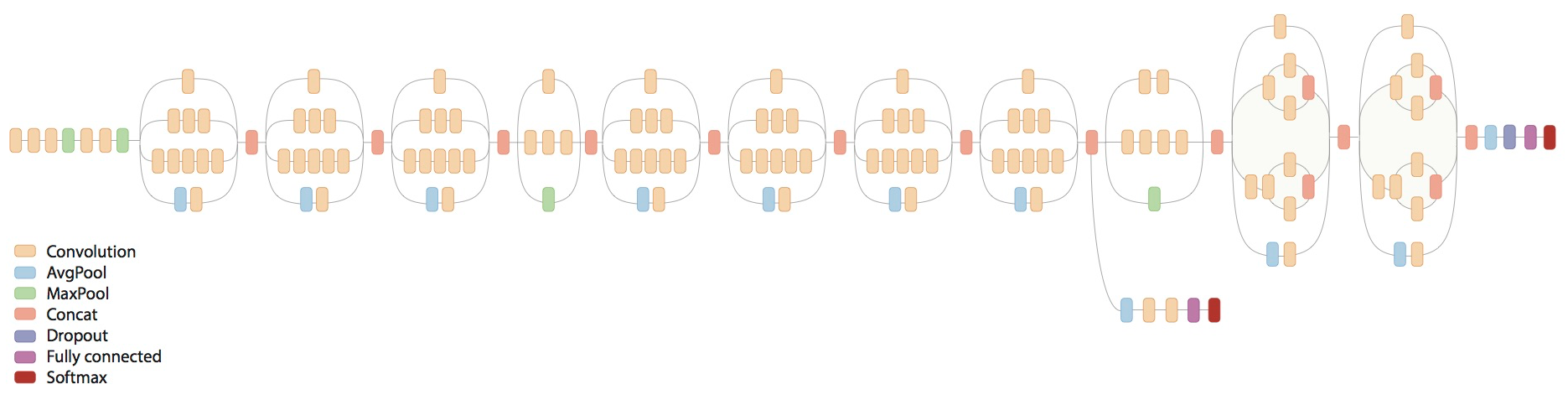}
\caption{Architecture of Inception v3}
\label{fig:inc}       % Give a unique label
\end{figure}

ImageNet dataset is randomly divided into training set (1,009,124 datapoints) and testing set (252,281 datapoints) with random guess 0.1\% due to 1,000 classes.

\subsection{Tuning parameter}

For training linear SVM models, it needs to tune the positive constant $C$ in SVM algorithms for keeping the trade-off between the margin size and the errors. We use the cross-validation (hold-out) protocol to find-out the best value $C=100,000$. LIBLINEAR uses L2-regularized Logistic Regression that is very closed to the softmax classifier used in deep learning networks, e.g. Inception v3 \cite{SzegedyVISW_inception_15}.

Training dataset is split into 8 blocks (the block size  127,000 requires about 2GB RAM) for inc-$k$SGD learning. The parameter $k$ local SGD models (number of clusters) of $k$SGD is set to $300$ so that each cluster has about $500$ datapoints. The idea gives a trade-off between the generalization capacity and the computational cost. Furthermore, the number of epochs and learning rate $eta$ of SGD are 50 and 0.001, respectively. 

Due to the Raspberry Pi 4 (Broadcom BCM2711, Quad core Cortex-A72) and the PC (Intel(R) Core i7-4790 CPU, 4 cores) used in the experimental setup, the number of threads is setting to 4 for all training tasks.

\subsection{Classification results}
We obtain classification results of Inc$k$lSGD, full SGD and LIBLINEAR in Table \ref{tab:res}, Fig. \ref{fig:time} and Fig. \ref{fig:acc}. The fastest training algorithm is in bold-faced and the second one is in italic. The same presentation format is accorded to performance in terms of classification accuracy, demanded memory size.
%We obtain classification results of Inc-$k$SGD, full SGD and LIBLINEAR in table \ref{tab:res}. The fastest training algorithm is in bold-faced and the second one is in italic. The same presentation format is used for classification accuracy.

Given the differences in implementation, including the programming language (C++ versus Python), computer (PC Intel(R) Core i7-4790 CPU, 4 cores, 32 GB RAM versus Raspberry Pi 4 Broadcom BCM2711, Quad core Cortex-A72, 4GB RAM), the comparison of training time is not really fair. But our Inc-$k$SGD achieves interesting results.

\begin{table}
\caption{Overall classification accuracy}
\label{tab:res}
\centering
\resizebox{.85\linewidth}{!}{
\begin{tabular}{|c|l|c|c|c|c|c|}
\hline	
 No & Algorithm & Language & Machine & Demanded memsize (GB) & Time (min) & Accuracy (\%) \\
\hline
1 & Inc-$k$SGD & Python & RPi4 & \textbf{2} & \textit{129.48} & \textbf{75.61} \\
\hline
2 & Full-SGD & Python  & PC & \textit{30 }& \textbf{179.29} & \textit{74.45} \\
\hline
3 & LIBLINEAR & C/C++ & PC & \textit{30} & 9,813.58 & 73.66 \\
\hline
\end{tabular}
} % resize
\end{table}

Our Inc-$k$SGD classifies ImageNet dataset in $129.48$ minutes with $75.61$\% accuracy. The full SGD achieves $74.45$\% accuracy with $179.29$ minutes in the training time. LIBLINEAR takes $9,813.58$ minutes for training the classification model with $73.66$\% accuracy. 

In the comparison of training time among algorithms, we can see that the Inc-$k$SGD with the Raspberry Pi is fastest training algorithm. Our Inc-$k$SGD with the Raspberry Pi is $75.79$ times faster than LIBLINEAR with the PC. The full SGD is $54.74$ times faster than LIBLINEAR. The full SGD on the PC is
$1.38$ times longer than the Inc-$k$SGD on the Raspberry Pi. 

In terms of overall accuracy, the Inc-$k$SGD gives the highest accuracy in the classification. The comparison, algorithm by algorithm, shows that the superiority of Inc-$k$SGD on LIBLINEAR corresponds to $1.95$\%. Inc-$k$SGD also improves $1.16$\% compared to the full SGD. 

Our Inc-$k$SGD training algorithm requires about 2 GB RAM against at least 30 GB RAM being used by the full SGD and LIBLINEAR. 

The classification results show that our Inc-$k$SGD algorithm is efficient for handling such large-scale multi-class datasets with the Raspberry Pi.

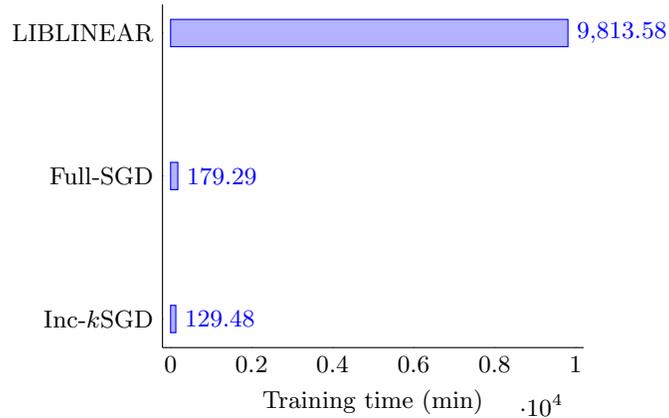
\begin{figure}
\centering
\resizebox{.75\linewidth}{!}{
\begin{tikzpicture}
  \begin{axis}[
    height=6cm, width=7cm,    
    xbar,
    tickwidth         = 0pt,
    ytick             = data,
    axis y line*=left,
    axis x line*=bottom,    
    xmin=0, xmax=10000,    
    enlarge y limits  = 0.1,
    enlarge x limits  = 0.02,
    symbolic y coords = {Inc-$k$SGD, Full-SGD, LIBLINEAR},
	nodes near coords, nodes near coords align={horizontal},
    xlabel = {Training time (min)},
    %legend pos=south east,
  ]
  
  \addplot coordinates {
	(129.48,Inc-$k$SGD) 
	(179.29,Full-SGD)                    
	(9813.58,LIBLINEAR) 
  };
  %\legend{SVM}
  \end{axis}
\end{tikzpicture}
                  
} % resize  
\caption{Training time (min)}
\label{fig:time}       % Give a unique label
\end{figure}

\begin{figure}
\centering
\resizebox{.75\linewidth}{!}{
\begin{tikzpicture}
  \begin{axis}[
    height=6cm, width=7cm,    
    xbar,
    tickwidth         = 0pt,
    ytick             = data,
    axis y line*=left,
    axis x line*=bottom,    
    xmin=60, xmax=80,    
    enlarge y limits  = 0.1,
    enlarge x limits  = 0.02,
    symbolic y coords = {Inc-$k$SGD, Full-SGD, LIBLINEAR},
	nodes near coords, nodes near coords align={horizontal},
    xlabel = {Accuracy (\%)},
    %legend pos=south east,
  ]
  
  \addplot coordinates {
	(75.61,Inc-$k$SGD) 
	(74.45,Full-SGD)                    
	(73.66,LIBLINEAR) 
  };
  %\legend{SVM}
  \end{axis}
\end{tikzpicture}
                  
} % resize  
\caption{Overall classification accuracy}
\label{fig:acc}       % Give a unique label
\end{figure}
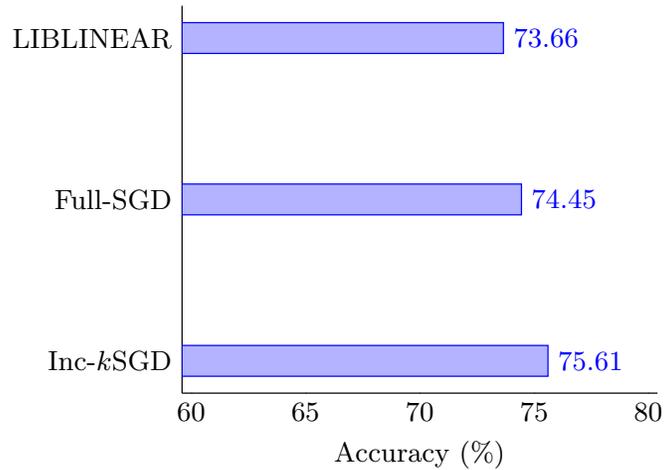

\section{Conclusion and future works}
\label{sec:conclusion}

We have presented the new incremental local SGD (Inc-$k$SGD) algorithm tailored on the Raspberry Pi to handle ImageNet challenging ILSVRC 2010 dataset having 1,261,405 images with 1,000 classes. The Inc-$k$SGD trains an ensemble of local SGD models by sequentially loading small data blocks for learning local SGD models. The local SGD uses $k$means algorithm to split the data block into $k$ partitions and then it learns in the parallel way SGD models in each data partition to classify the data locally. The numerical test results on ImageNet challenging dataset show that our Inc-$k$SGD algorithm with the Raspberry Pi 4 is $75.79$ times faster than the state-of-the-art LIBLINEAR on the PC with an improvement of $1.95$\% accuracy.

In the near future, we will develop the distributed implementation for the incremental local SGD algorithm on an in-memory cluster-computing platform with the Raspberry Pis. 
\subsubsection{Acknowledgments}
\label{sec:Acknowledgments}
This work has received support from the College of Information Technology, Can Tho University. We would like to thank very much the Big Data and Mobile Computing Laboratory.
\bibliographystyle{splncs04}
\bibliography{refs}

\end{document}